\documentclass{article}

\usepackage{arxiv}

\usepackage[utf8]{inputenc} 
\usepackage[T1]{fontenc}    
\usepackage{hyperref}       
\usepackage{url}            
\usepackage{booktabs}       
\usepackage{amsfonts}       
\usepackage{nicefrac}       
\usepackage{microtype}      
\usepackage{lipsum}
\usepackage{graphicx}

\usepackage{amsmath,amssymb,graphicx,booktabs}
\usepackage{acronym}
\usepackage{booktabs}
\usepackage[disable]{todonotes}	
\usepackage{amssymb}
\usepackage{amsthm}
\usepackage{subcaption}
\usepackage{dblfloatfix}
\usepackage{tikz}
\usetikzlibrary{decorations.pathreplacing}
\usetikzlibrary{fadings}
\usepackage{pifont} 
\newcommand{\cmark}{\ding{51}}
\newcommand{\xmark}{\ding{55}}
\usepackage[inline]{enumitem}

\usepackage[
	backend=biber,
	style=authoryear,
	sorting=nyt,
	maxbibnames=7,
      maxcitenames=1
	]{biblatex}
\DeclareSourcemap{ 
  \maps[datatype=bibtex]{
    \map[overwrite=true]{
      \step[fieldset=urldate, null]
    }
  }
}

\AtEveryBibitem{\clearfield{month}} 

\title{\textsc{EviNAM}: Intelligibility and Uncertainty via Evidential Neural Additive Models}

\author{
    Sören Schleibaum \\
    Clausthal University of Technology \\
    \texttt{soeren.schleibaum@tu-clausthal.de} \\
    \And
    Anton Frederik Thielmann \\
    Amazon Music \\
    \And
    Julian Teusch \\
    Clausthal University of Technology \\
    \texttt{julian.teusch@tu-clausthal.de} \\
    \And
    Benjamin Säfken \\
    Clausthal University of Technology \\
    \texttt{benjamin.saefken@tu-clausthal.de} \\
    \And
    Jörg P. Müller \\
    Clausthal University of Technology \\
    \texttt{joerg.mueller@tu-clausthal.de} \\
}

\begin{document}

\newacro{BNN}[BNN]{Bayesian Neural Network}
\newacro{CRPS}[CRPS]{Continuous Ranked Probability Score}
\newacro{DEC}[DEC]{Deep Evidential Classification}
\newacro{DER}[DER]{Deep Evidential Regression}
\newacro{GAM}[GAM]{Generalized Additive Model}
\newacroplural{GAM}[GAMs]{Generalized Additive Models}
\newacro{NAM}[NAM]{Neural Additive Model}
\newacroplural{NAM}[NAMs]{Neural Additive Models}
\newacro{NAMLSS}[NAMLSS]{NAM for Location, Scale, and Shape}
\newacro{MAE}[MAE]{Mean Absolute Error}
\newacro{MLP}[MLP]{Multilayer Perceptron}
\newacro{NIG}[NIG]{Normal Inverse Gamma distribution}
\newacro{NLL}[NLL]{Negative Log-Likelihood}
\newacro{SHAP}[SHAP]{Shapley Additive Explanations}

\newcommand\evinam{{\textsc{EviNAM}}}

\maketitle
\begin{abstract}
  Intelligibility and accurate uncertainty estimation are crucial for reliable decision-making. In this paper, we propose {\evinam}, an extension of evidential learning that integrates the interpretability of Neural Additive Models (NAMs) with principled uncertainty estimation. Unlike standard Bayesian neural networks and previous evidential methods, {\evinam} enables, in a single pass, both the estimation of the aleatoric and epistemic uncertainty as well as explicit feature contributions. Experiments on synthetic and real data demonstrate that {\evinam} matches state-of-the-art predictive performance. While we focus on regression, our method extends naturally to classification and generalized additive models, offering a path toward more intelligible and trustworthy predictions.
\end{abstract}


\section{Introduction}

While the effectiveness of neural networks in various tasks is astonishing, they are \emph{black boxes}. This is caused by (1) \emph{nonlinearity}, (2) \emph{unexpected behavior} that can make human users curious~\parencite{molnar.2025}, and (3) \emph{information representation} through a relatively high number of trainable, interacting parameters that differs from the mental models of humans~\parencite{zhang.2021c}.

One attempt to ``open'' such black boxes is to apply an explainable artificial intelligence method post hoc, like \emph{LIME}~\parencite{ribeiro.2016} or \emph{SHAP}~\parencite{lundberg.2017}. Both methods estimate the contribution of a feature by perturbing the input and using the resulting changes to fit a local surrogate model, whose weights correspond to the feature's contribution estimate.

An alternative are directly intelligible models, like linear regression or \emph{\acp{GAM}}. \textcite{hastie.2017} proposed the latter to capture complex, nonlinear relations between the input and a target. \acp{GAM} learn a function $f$ parameterized by $\theta$ per feature $j \in J$ and sum their outputs to form the prediction:
\begin{equation}
  \mathbb{E}(y_i) = \phi \left( b + \sum\nolimits_{j=1}^{J} f_{\theta^j}(x_i^j) \right),
\end{equation}
where $y_i \in \mathbb{R}$, $\phi(\cdot)$ is a link function, e.g., the identity or $\log$ function, and $b$ is the global intercept for all samples.
A promising enhancement of~\parencite{hastie.2017} was proposed by \textcite{agarwal.2021}; the authors trained one neural network per feature (rather than, e.g., LOWESS~\parencite{cleveland.1979}) to learn a function per feature. Such \acp{NAM} can offer strong predictive performance, handle large datasets, and explicitly reveal all feature contributions to the prediction.

While modeling feature contributions offers intelligibility to explainees, by default, neural networks neither enable the identification of out-of-domain samples nor do they expose confidence ~\parencite{amini.2020,gawlikowski.2023}. However, both abilities can help explainees to harmonize their mental model with the behavior learned by a neural network, reducing the unexpected behavior and enriching the information available to them.

Therefore, approaches that provide these abilities by explicitly modeling uncertainty like \emph{\ac{DER}} -- proposed by \textcite{amini.2020} -- are promising. Besides the mean response $\mathbb{E}(y_i)$, these approaches enable to derive the \emph{aleatoric} and \emph{epistemic uncertainty}; while the former refers to uncertainty inherent in the data, e.g., when rain blurs the sensors of an autonomous vehicle detecting a pedestrian, the latter captures uncertainty in a model’s prediction, e.g., a clear but unfamiliar driving scene. Learning both types of uncertainty contrasts with most existing approaches to uncertainty modeling \parencite{hullermeier.2021}. Unlike sampling-dependent \acp{BNN} methods, \ac{DER} enables explainees to validate predictions using uncertainty estimates in a single pass. Evidential learning is criticized for struggling to quantify epistemic uncertainty -- a non-data property \parencite{bengs.2022,meinert.2023} -- yet it remains valuable when decisions can be handed off to humans under high epistemic uncertainty \parencite{meinert.2023}.

Under the same additive constraint as \parencite{hastie.2017,agarwal.2021}, several follow-ups improved predictive performance \parencite{thielmann.2024a,enouen.2022}, including tree- and NODE-based shape functions \parencite{nori.2019,chang.2022} and higher-order interactions \parencite{kim.2022b,thielmann.2025}. Most additive models target only the mean \parencite{kneib.2023}; distributional variants such as \ac{NAMLSS} \parencite{thielmann.2024} and NODE-GAMLSS \parencite{de.2024} relax this. For uncertainty, ensembles quantify disagreement \parencite{lakshminarayanan.2017} and \acp{BNN} treat parameters as random variables~\parencite{gawlikowski.2023}. \parencite{bouchiat.2024} combined \acp{NAM} and \acp{BNN} to \emph{BNAM} to achieve additivity and epistemic uncertainty estimates. While the training effort is high for ensemble-based methods and \acp{BNN} require sampling, evidential learning methods like \ac{DER} \parencite{amini.2020} -- refined by \textcite{meinert.2023} -- yield aleatoric and epistemic uncertainty in $\mathcal{O}(1)$. Yet none simultaneously offers intelligibility through additivity and both uncertainties in a single pass; see Table~\ref{tab:evinam_vs_related_work}.

\begin{table}[h!]
  \centering
  \caption{Comparison of the proposed method {\evinam} to the related work.}
  \begin{tabular}{lcccc}
    \toprule
    Method             & Additivity & $u_{\text{al}}$ & $u_{\text{ep}}$ & $u$ at inference \\
    \midrule
    \ac{BNN}           & \xmark     & \cmark          & \cmark          & \xmark           \\
    BNAM               & \cmark     & \xmark          & \cmark          & \xmark           \\
    \ac{DER}           & \xmark     & \cmark          & \cmark          & \cmark           \\
    \ac{NAM}           & \cmark     & \xmark          & \xmark          & \xmark           \\
    \ac{NAMLSS}        & \cmark     & \cmark          & \xmark          & \cmark           \\
    \textbf{{\evinam}} & \cmark     & \cmark          & \cmark          & \cmark           \\
    \bottomrule
    \label{tab:evinam_vs_related_work}
  \end{tabular}
\end{table}

Thus, we present a novel approach named {\evinam} (\underline{Evi}dential \underline{N}eural \underline{A}dditive \underline{M}odel) for both regression and classification. Our extension of evidential learning can -- unlike \ac{BNN}-based approaches and vanilla \ac{DER} -- make the aleatoric and epistemic uncertainty as well as the contribution of each feature to a prediction explicit, all in a single pass.

\section{Problem Definition}
\label{sec:ProblemDefinition}
Consider a dataset $ \mathcal{D} = {\{ (x_{i}, y_{i}) \}}_{i=1}^N $ of $ N $ paired samples with $ x_i \in \mathbb{R}^J $ and $ y_i \in \mathbb{R} $ making it a regression problem. We assume that the targets $ \{ y_1, \ldots , y_N \} $ are drawn from a Gaussian. Further, we assume that the mean and variance of these distributions are unknown. We aim to learn a function $f_\theta$ parameterized by $\theta$ that provides the following information for an unseen sample $x_i$: The (1) actual prediction $ \gamma_i $, the (2) aleatoric uncertainty $ u_{\text{al},i} $, the (3) epistemic uncertainty $ u_{\text{ep},i} $, and the (4) contribution of each feature $ j \in J $ to the prediction $f_{\theta^j}^{\gamma}(x_i^j)$. Thereby, we intend to make the model's decision as well as its confidence transparent to human decision-makers, enable them to identify out-of-domain samples at inference time.

\section{Approach}
\label{sec:approach}
We first outline \ac{DER}, then integrate \ac{DER} and \ac{NAM}. We motivate forwarding the nonlinearity to enforce distributional constraints (e.g., $\sigma^2 > 0$). Finally, although we focus on regression, we show {\evinam} also applies to classification.”

\paragraph{Deep Evidential Regression}
\textcite{amini.2020} assumed that the target is Gaussian-distributed and both $\mu$ and $\sigma^2$ of the Gaussian are unknown. Therefore, they modeled the problem as follows:
\begin{equation}
  ( y_1, \cdots , y_N) \sim \mathcal{N}(\mu, \sigma^2) \qquad
  \mu                  \sim \mathcal{N}(\gamma, \sigma^2 v^{-1}) \qquad
  \sigma^2             \sim \Gamma^{-1} (\alpha, \beta),
\end{equation}
where $ \Gamma^{-1} $ is the inverse Gamma distribution. This induces a \ac{NIG} prior with parameters $ \boldsymbol{m}_i = (\gamma_i, \upsilon_i, \alpha_i, \beta_i) $ for a sample $i$. The distributional parameters must satisfy certain requirements: $ \gamma_i \in \mathbb{R}, \, \upsilon_i > 0, \, \alpha_i > 1, \, \beta_i > 0 $. The first distributional parameter is the mean prediction: $ \mathbb{E}[\mu] = \gamma_i $. \textcite{amini.2020} proposed to learn function mappings $f_{\theta^j}^{k}$ represented by a neural network with a set of weights $ \theta^j $ per feature $j$ and distributional parameter $k \in \{ \gamma, \upsilon, \alpha, \beta \} $. 

\textcite{meinert.2023} refined \parencite{amini.2020} by proposing the usage of the following loss function:
\begin{equation}\label{eq:MeinertLoss}
  \mathcal{L}_i (\theta) = - \log L_i^{\text{NIG}}(\theta) + \lambda \left| \frac{y_i - \gamma_i}{w_{\text{St},i}} \right|^p \Phi_i, \quad \text{with} \quad w_{\text{St},i} = \sqrt{ \frac{\beta_i (1 + \upsilon_i)}{\alpha_i \upsilon_i} }; \quad
  \Phi_i   = 2 \upsilon_i + \alpha_i.
\end{equation}
$\lambda > 0$ is a regularization coefficient and $p > 0$. Further, the authors argued for an alternative calculation of the aleatoric and epistemic uncertainty from the parameters of the learned \ac{NIG}:
\begin{equation} \label{eq:UncertaintiesMeinert23}
  u_{\text{al},i} = \mathbb{E}[\sigma_i] = w_{\text{St},i};  \qquad
  u_{\text{ep},i} = \upsilon_i^{-\frac{1}{2}}.
\end{equation}
They proposed this extension to overcome an overparameterization-issue in the loss from \parencite{amini.2020}.

\paragraph{Integrating DER and NAM} \textcite{agarwal.2021} proposed to learn one neural network $f$ per feature $j$ and to sum their outputs to receive a prediction:
\begin{equation} \label{eq:NAM}
  \mathbb{E}(y_i) = \phi \left( b + \sum\nolimits_{j=1}^{J} f_{\theta^j}(x_i^j) \right),
\end{equation}
where $y_i \in \mathbb{R}$, $\phi(\cdot)$ is a link function, e.g., the identity or $\log$ function, and $b$ is the global intercept for all samples. To integrate \ac{NAM} and \ac{DER}, we propose to learn each \ac{NIG} parameter $k$ analogous to Equation~\ref{eq:NAM}:
\begin{equation} \label{eq:DERandNAM}
  k_i = \phi^{k} \left( b^{k} + \sum\nolimits_{j=1}^{J} f_{\theta^j}^{k} (x_i^j) \right).
\end{equation}
Each feature network $ f_{\theta^j}^{k}: \mathbb{R} \rightarrow \mathbb{R}$ is parameterized by $\theta^j \in \mathbb{R}^P $.

\paragraph{Forwarding Nonlinearity} Often, distributional parameters must fulfill specific requirements, e.g., $\sigma^2 > 0$ when learning a Gaussian. Therefore, \textcite{amini.2020,meinert.2023,thielmann.2024} used nonlinear link functions $\phi(\cdot)$, e.g., the softplus function for $\upsilon$: $ \phi^{\upsilon} (z) = \log ( 1 + e^z ) $. The consequence is that the additivity can be violated because:
\begin{equation}
  \exists (f_{\theta^n}^k(x_i^n), f_{\theta^m}^k(x_i^m)) \in \mathbb{R}^2:\ \phi^k \left(f_{\theta^n}^k(x_i^n) + f_{\theta^m}^k(x_i^m)\right) \neq \phi^k \left(f_{\theta^n}^k(x_i^n)\right) + \phi^k \left(f_{\theta^m}^k(x_i^m)\right),
\end{equation}
for $n, m \in J, n \neq m$. Thus, the contribution of a feature $j$ to the distributional parameters $k$ is obscured when $\phi^k(.)$ is nonlinear. We propose to forward the nonlinearity to the feature-level for each distributional parameter by replacing Equation~\ref{eq:DERandNAM} with:
\begin{equation} \label{eq:proposed}
  k_i = \phi^{k} \left(b^{k} \right) + \sum\nolimits_{j=1}^{J} \phi^{k} \left( f_{\theta^j}^{k}(x_i^j) \right),
\end{equation}
to make the contribution of each feature to each distributional parameter explicit while preserving both the additivity (important for the intelligibility) and the distribution's requirements.

Usually, for \ac{NIG}s, $ \phi^\gamma $ is simply the identity function, $ \phi^\beta(\cdot) = \phi^\upsilon(\cdot) = \log ( 1 + e^{(\cdot)} )$, and $ \phi^\alpha(\cdot)=\phi^\upsilon(\cdot) + 1 $. Alternative functions for $ \phi^{k} $ are possible; our approach of forwarding the nonlinearity can be easily adapted to accommodate other potential activation functions.

\paragraph{Applicability to Classification} \textcite{sensoy.2018} introduced \ac{DEC}, an approach that learns a Dirichlet distribution $\boldsymbol{p}_i \sim \mathrm{Dir}(\alpha_i^1, \dots, \alpha_i^C)$ for a sample $i$, where $C$ refers to the number of classes and $\alpha_i^n \ge 1,\ \forall n \in \{1,\dots,C\}$. The class probability can be computed via $\hat{p}_i^c = \tfrac{\alpha_i^c}{S_i}$, where $S_i=\sum_{n=1}^{C} \alpha_i^n$; aleatoric and epistemic uncertainty follow from the Dirichlet. Similar to regression, we propose to forward the nonlinearity used to ensure that the distributional parameters fulfill certain criteria, here $\alpha_i^n \ge 1$. In \ac{DEC}, $\alpha_i^c = 1 + e_i^c$ with $e_i^c$ being computed via a neural network. Integrating \ac{NAM} with \ac{DEC} and forwarding the nonlinearity yields:
\begin{equation}
  \alpha_i^c = 1 + \sum\nolimits_{j=1}^{J} \phi^{\alpha^c} \left( e_{i}^{c,j} \right),
\end{equation}
where $j$ refers to the feature and $\phi(\cdot)$ ensures positive values, e.g., via softplus. Only by forwarding the nonlinearity do feature contributions to class probabilities remain additive.

Typically, with additive models (e.g., \ac{NAM}) for classification, $\mathrm{softmax}(\cdot)$ is used to map logits to probabilities. In contrast to {\evinam}, the numerator of the $\mathrm{softmax}(\cdot)$ function violates additivity ($\exists (a, b) \in \mathbb{R}^2:\ e^{a + b} \neq e^a + e^b$). This issue can be tackled by using another link function or by forwarding the constraint $\sum\nolimits_{c=1}^{C} p_i^c = 1$ to the feature level!

\section{Computational Experiments}
Here, we
\begin{enumerate*}[label=(\arabic*)]
  \item analyze the proposed forwarding of the nonlinearity,
  \item estimate uncertainties on synthetic data,
  \item compare the predictive performance of {\evinam} against several baseline methods,
  \item assess its computational complexity, and
  \item illustrate its intelligibility on a real-world dataset.
\end{enumerate*}

\paragraph{Forwarding Nonlinearity} We compare the effect of the proposed forwarding of nonlinearity on the predictive performance. We use the first five datasets from the OpenML suite 353~\parencite{fischer.2023} and compare two alternatives for the architecture via average rank after hyperparameter tuning on the validation data:
\begin{enumerate*}[label=(A\arabic*)]
  \item \emph{Nonlinearity at feature-level} -- the proposed alternative -- versus
  \item \emph{nonlinearity at sum} -- the baseline.
\end{enumerate*}
We evaluate three metrics: \ac{NLL}, \ac{CRPS}, and \ac{MAE}. Lower rank is better; with two variants, 1 is best and 2 worst. The proposed alternative achieved an average rank of 1.6, \textbf{1.4}, and \textbf{1.4} for the three metrics compared to \textbf{1.4}, 1.6, and 1.6. As the proposed alternative performs on par and its advantages regarding intelligibility, we select A1 for the following experiments.

\begin{figure*}[t]
  \centering
  \begin{minipage}[t]{0.24\textwidth}
    \centering
    \includegraphics[width=\linewidth]{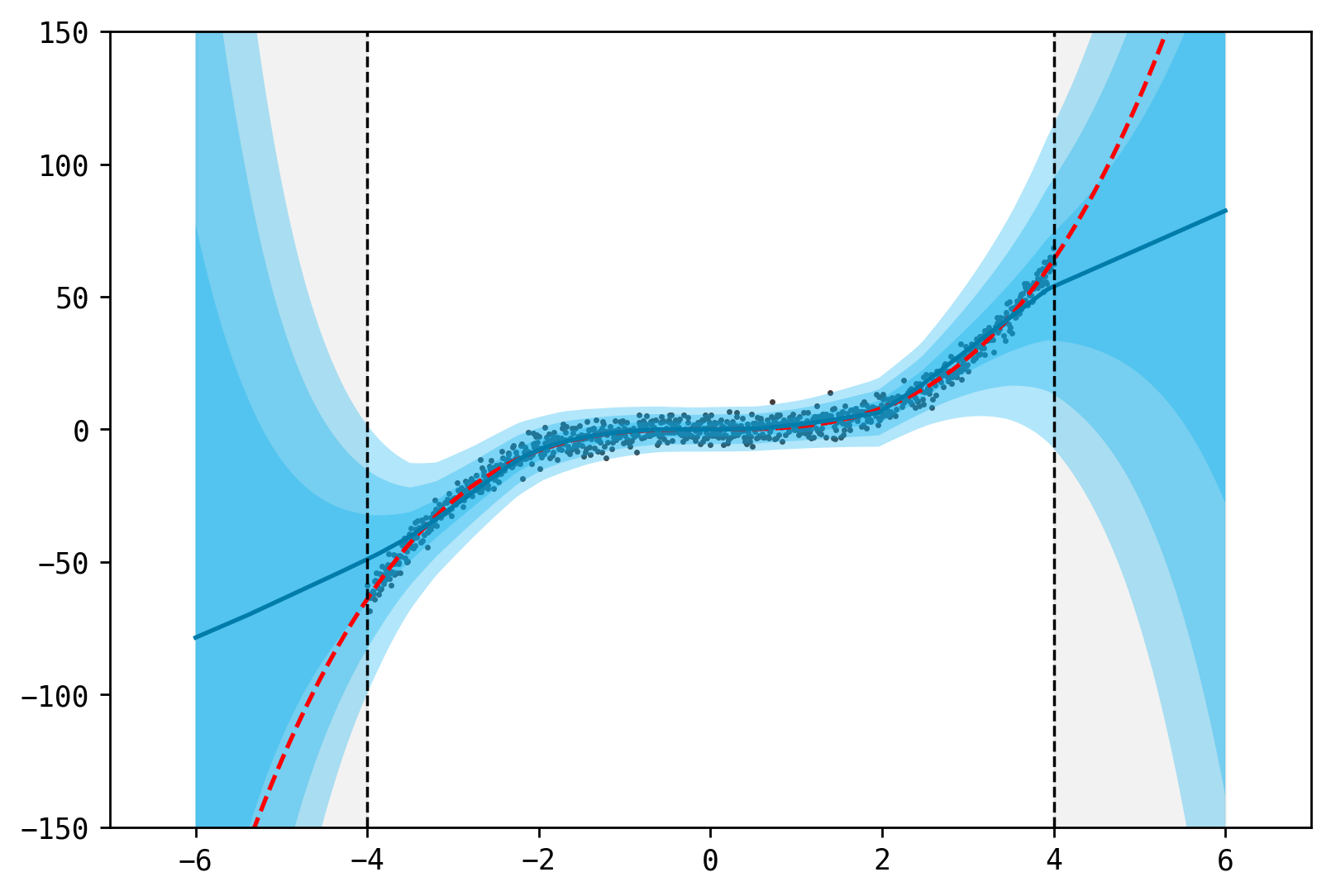}
    \caption*{(a) $\text{DER}_{\text{MLP}}$}
  \end{minipage}
  \begin{minipage}[t]{0.24\textwidth}
    \centering
    \includegraphics[width=\linewidth]{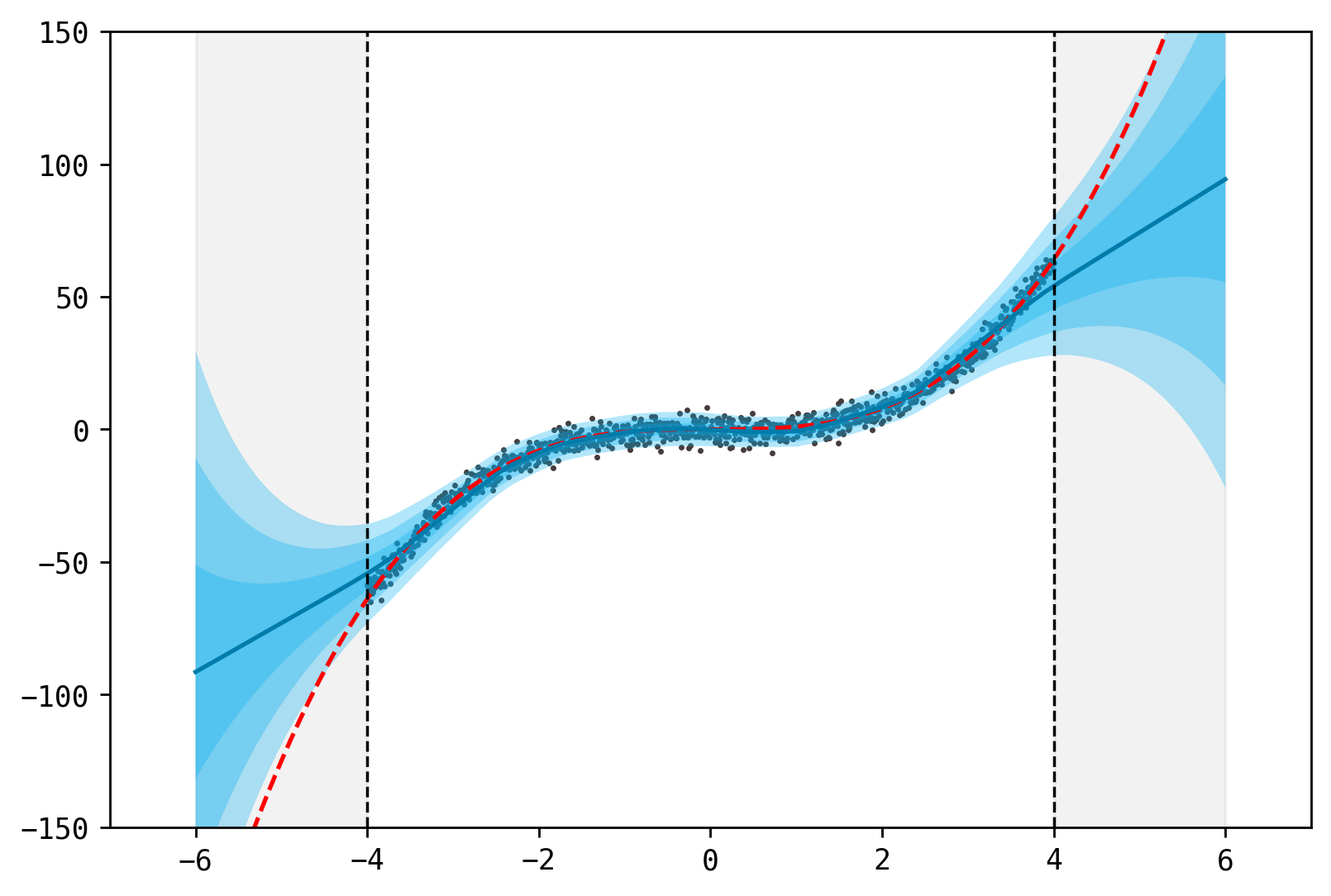}
    \caption*{(b) {\evinam}}
  \end{minipage}
  \begin{minipage}[t]{0.24\textwidth}
    \centering
    \includegraphics[width=\linewidth]{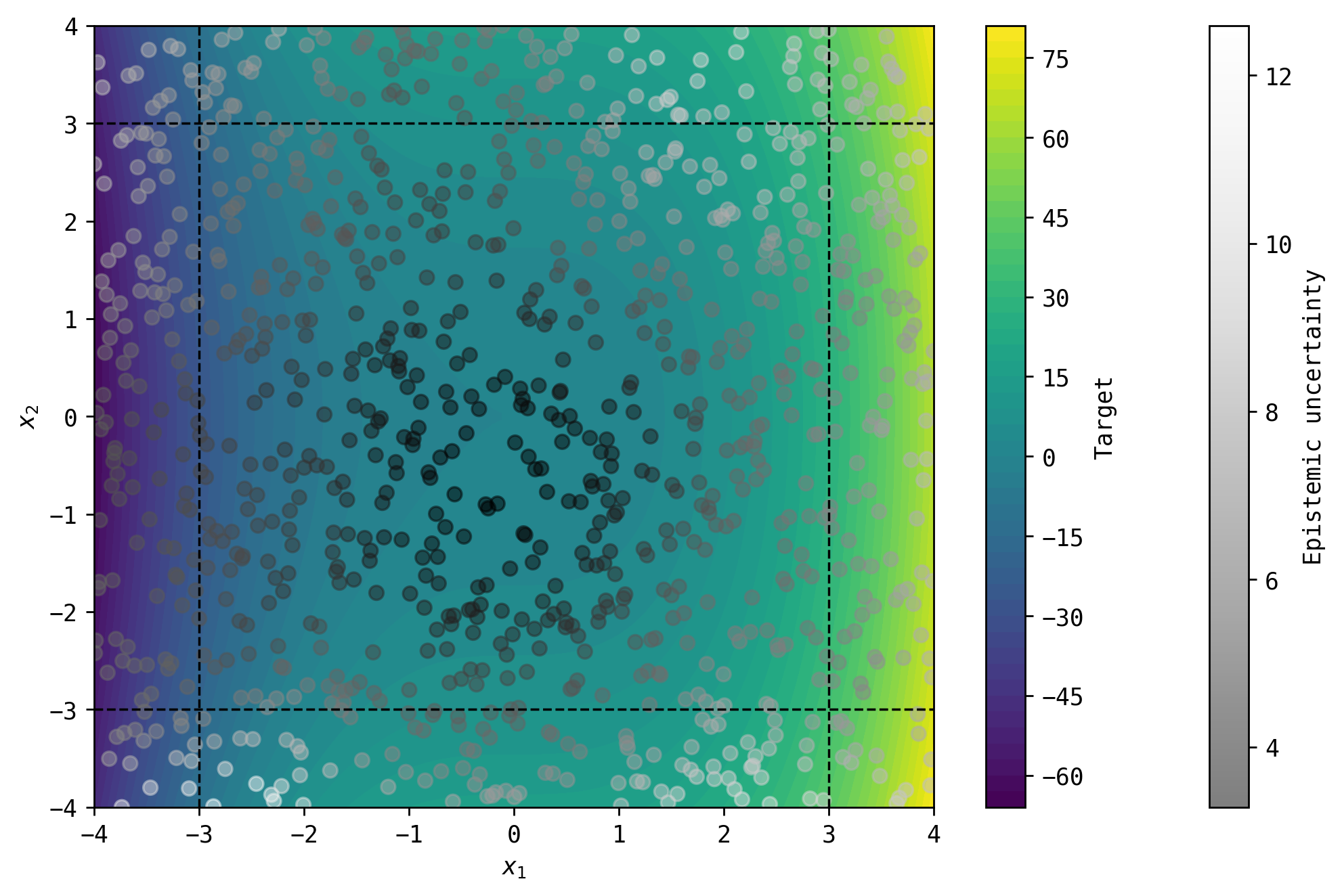}
    \caption*{(c) $\text{DER}_{\text{MLP}}$}
  \end{minipage}
  \begin{minipage}[t]{0.24\textwidth}
    \centering
    \includegraphics[width=\linewidth]{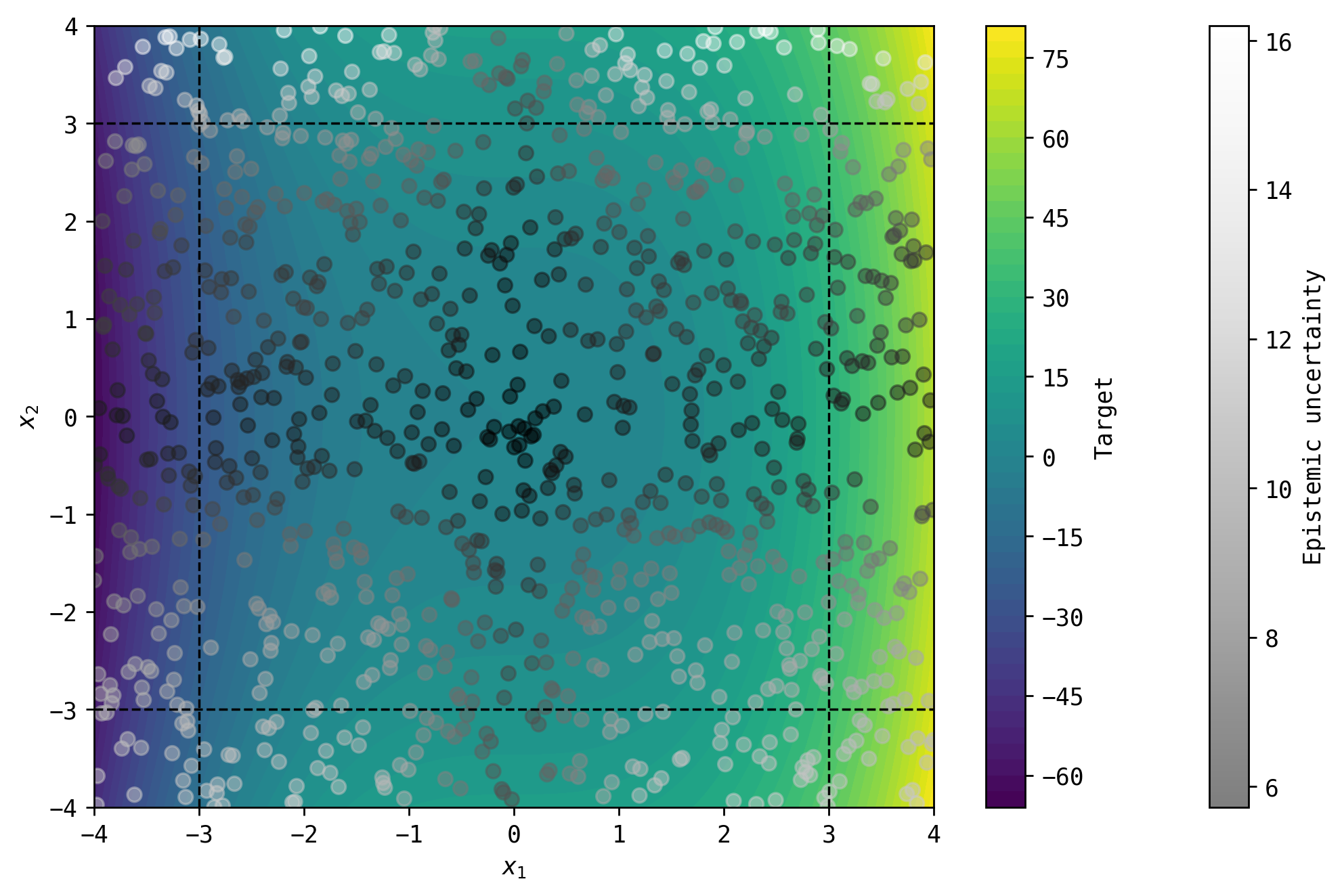}
    \caption*{(d) {\evinam}}
  \end{minipage}
  \caption{(a) and (b) show the epistemic uncertainty on a 1D cubic dataset; white and gray mark training and test regions, red is the ground truth, black the prediction, and blue shadows the epistemic uncertainty. (c) and (d) show epistemic uncertainty estimation on a 2D cubic dataset via a gray scale.}
  \label{fig:EpistemicUncertaintyOnCubic}
\end{figure*}

\paragraph{Epistemic Uncertainty Estimation on Synthetic Data} We compared the uncertainty estimation of {\evinam} to that of vanilla \ac{DER} with a \ac{MLP} ($\text{DER}_{\text{MLP}}$) on synthetic data. For $\text{DER}_{\text{MLP}}$ we chose the refined version from \parencite{meinert.2023}. Following \parencite{amini.2020}, we use 1D dataset: $ y = x^3 + \epsilon $ with $ \epsilon \sim \mathcal{N} (0, 3)$. As shown in Figure~\ref{fig:EpistemicUncertaintyOnCubic}, {\evinam} captures a similar epistemic pattern as $\text{DER}_{\text{MLP}}$.

As the architecture of {\evinam} and $\text{DER}_{\text{MLP}}$ is similar when using 1D data, we constructed another 2D dataset: $ y = ({x^{(1)}}^3 + \epsilon^{(1)}) + ({x^{(2)}}^2 + \epsilon^{(2)}$) with $ \epsilon^{(1)} \sim \mathcal{N} (0, 5)$ and $ \epsilon^{(2)} \sim \mathcal{N} (0, 1)$. When we used 1000 training samples in $[-3, 3]$ and 1000 test samples in $[-4, 4]$, {\evinam} (on average $R^2=0.9210$) performed similarly to $ \text{DER}_{\text{MLP}} $ ($R^2=0.9429$). We show the learned epistemic uncertainty $u_{\text{ep},i}$ in Figure~\ref{fig:EpistemicUncertaintyOnCubic}. As expected, both models learned a higher $u_{\text{ep},i}$ at higher values of $x^{(1)}$ and $x^{(2)}$; the $u_{\text{ep},i}$ depended more on $x^{(1)}$. As expected, the $u_{\text{ep},i}$ was lower in the center ($|x^{(i)}| < 3 \quad \forall i \in \{1,2\}$) where the inputs lie within the training domain.

\paragraph{Predictive Performance} We compared the predictive performance of {\evinam} on 24 real-world datasets from the OpenML suite 353. We used the provided split to separate 10\% for testing and split the remaining 90\% into a 72\% train and 18\% validation. We applied Z-score normalization to the numerical features and the target, and one-hot-encoded categorical features. We selected \ac{NAM}, \ac{NAMLSS}, and an ensemble of \acp{NAM} (\emph{EnsNAM}) as baselines, because they also support per-feature contribution extraction. We trained with the Adam optimizer and a learning-rate scheduler for up to 5000 epochs using early stopping ($\text{patience}=50$) and tuned hyperparameters via Bayesian optimization (25 trials). Results are averaged over 10 runs on the test data.

Following prior work, we used three metrics to compare the predictive performance:
\begin{enumerate*}[label=(\arabic*)]
  \item \ac{MAE},
  \item \ac{NLL}, and
  \item \ac{CRPS}.
\end{enumerate*}
Further, we computed the average rank achieved per model and metric and compared the differences between the methods via the nonparametric Friedman test (significance level $\alpha=0.05$) with the Nemenyi post hoc procedure -- see~\cite{derrac.2011}.

As shown in Table~\ref{tab:predictive_performance_regression}, {\evinam} outperformed \ac{NAM} and EnsNAM in terms of \ac{NLL} (\ac{NAM} with the actual variance) and \ac{CRPS} on most datasets, while falling slightly behind \ac{NAMLSS}. The latter is expected as {\evinam} learns more distributional parameters (4 for a \ac{NIG} vs. 2 for a Gaussian) which enable to derive both uncertainties. Surprisingly, {\evinam} was able to achieve the lowest \ac{MAE} for 5 datasets. These results are backed up by the average rank and the Friedman tests: For \ac{NLL}, {\evinam} trails \ac{NAMLSS}, matches \ac{NAM}, and beats EnsNAM; for \ac{CRPS}, {\evinam} and \ac{NAMLSS} tie and outperform \ac{NAM} and EnsNAM; for \ac{MAE}, differences are not significant.

\begin{table*}[t]
  \centering
  \caption{Regression results on test data via mean$\pm$standard deviation; average rank in last row.}
  \resizebox{\textwidth}{!}{
    \begin{tabular}{@{}lcccc|cccc|cccc@{}}
      \toprule
                                & \multicolumn{4}{c}{NLL $\downarrow$} & \multicolumn{4}{c}{CRPS $\downarrow$} & \multicolumn{4}{c}{MAE $\downarrow$}                                                                                                                                                                                                                                                     \\
      Dataset                   & NAM                                  & NAMLSS                                & EnsNAM                               & \textbf{{\evinam}}         & NAM                       & NAMLSS                    & EnsNAM           & \textbf{{\evinam}}        & NAM                       & NAMLSS                    & EnsNAM                    & \textbf{{\evinam}}        \\ \midrule
      Abalone                   & \underline{1.16$\pm$0.00}            & \textbf{1.10$\pm$0.07}                & 2.00$\pm$0.27                        & 1.59$\pm$0.60              & 0.40$\pm$0.00             & \textbf{0.36$\pm$0.01}    & 0.86$\pm$0.24    & \underline{0.36$\pm$0.00} & 0.50$\pm$0.01             & 0.49$\pm$0.01             & \underline{0.49$\pm$0.00} & \textbf{0.49$\pm$0.01}    \\
      Airfoil self noise        & 1.15$\pm$0.01                        & \textbf{0.52$\pm$0.03}                & 1.80$\pm$0.24                        & \underline{0.58$\pm$0.28}  & 0.40$\pm$0.01             & \textbf{0.32$\pm$0.00}    & 0.65$\pm$0.11    & \underline{0.35$\pm$0.07} & 0.53$\pm$0.01             & \textbf{0.48$\pm$0.00}    & 0.53$\pm$0.00             & \underline{0.51$\pm$0.09} \\
      Auction verification      & 1.05$\pm$0.00                        & \textbf{-0.68$\pm$0.07}               & 1.58$\pm$0.37                        & \underline{-0.12$\pm$0.25} & \textbf{0.33$\pm$0.00}    & \underline{0.38$\pm$0.00} & 0.55$\pm$0.21    & 0.41$\pm$0.03             & \underline{0.39$\pm$0.01} & 0.58$\pm$0.01             & \textbf{0.38$\pm$0.00}    & 0.60$\pm$0.03             \\
      Physiochemical protein    & \underline{1.22$\pm$0.00}            & \textbf{1.10$\pm$0.01}                & 2.52$\pm$0.41                        & 3.06$\pm$0.60              & \underline{0.44$\pm$0.00} & \textbf{0.43$\pm$0.00}    & 1.56$\pm$0.59    & 0.46$\pm$0.01             & \underline{0.61$\pm$0.00} & 0.62$\pm$0.01             & \textbf{0.61$\pm$0.00}    & 0.63$\pm$0.02             \\
      Superconductivity         & 0.99$\pm$0.00                        & \textbf{0.10$\pm$0.04}                & 2.31$\pm$0.94                        & \underline{0.66$\pm$0.13}  & 0.28$\pm$0.00             & \underline{0.21$\pm$0.00} & 1.04$\pm$0.84    & \textbf{0.18$\pm$0.00}    & \underline{0.25$\pm$0.00} & 0.30$\pm$0.00             & 0.30$\pm$0.00             & \textbf{0.25$\pm$0.00}    \\
      Naval propulsion plant    & 0.92$\pm$0.00                        & \textbf{-2.84$\pm$0.83}               & 6.34$\pm$0.59                        & \underline{-2.81$\pm$0.58} & 0.23$\pm$0.00             & \textbf{0.01$\pm$0.01}    & 104.12$\pm$43.70 & \underline{0.03$\pm$0.03} & 0.04$\pm$0.02             & \underline{0.02$\pm$0.01} & \textbf{0.02$\pm$0.01}    & 0.04$\pm$0.04             \\
      White wine                & 1.21$\pm$0.00                        & \textbf{1.15$\pm$0.01}                & 2.11$\pm$0.27                        & \underline{1.17$\pm$0.01}  & 0.44$\pm$0.00             & \textbf{0.43$\pm$0.00}    & 0.88$\pm$0.24    & \underline{0.44$\pm$0.00} & \underline{0.61$\pm$0.01} & 0.61$\pm$0.01             & \textbf{0.60$\pm$0.00}    & 0.61$\pm$0.01             \\
      Red wine                  & \underline{1.28$\pm$0.00}            & \textbf{1.21$\pm$0.02}                & 2.02$\pm$0.64                        & 1.32$\pm$0.03              & \underline{0.47$\pm$0.00} & \textbf{0.46$\pm$0.01}    & 0.61$\pm$0.06    & 0.48$\pm$0.01             & \underline{0.64$\pm$0.00} & 0.65$\pm$0.01             & \textbf{0.63$\pm$0.00}    & 0.66$\pm$0.02             \\
      Grid stability            & 1.04$\pm$0.00                        & \textbf{0.58$\pm$0.02}                & 2.37$\pm$1.83                        & \underline{0.71$\pm$0.14}  & 0.32$\pm$0.00             & \textbf{0.27$\pm$0.00}    & 0.35$\pm$0.03    & \underline{0.30$\pm$0.03} & \textbf{0.38$\pm$0.01}    & \underline{0.39$\pm$0.01} & 0.39$\pm$0.00             & 0.43$\pm$0.05             \\
      Video transcoding         & 1.14$\pm$0.00                        & \textbf{-0.62$\pm$0.09}               & 2.38$\pm$1.96                        & \underline{-0.41$\pm$0.20} & 0.36$\pm$0.00             & \underline{0.26$\pm$0.00} & 0.40$\pm$0.04    & \textbf{0.25$\pm$0.01}    & 0.41$\pm$0.01             & \textbf{0.38$\pm$0.00}    & 0.41$\pm$0.01             & \underline{0.39$\pm$0.02} \\
      Wave energy               & 0.92$\pm$0.00                        & \textbf{-3.03$\pm$0.31}               & 1.76$\pm$0.50                        & \underline{-2.42$\pm$3.69} & 0.23$\pm$0.00             & \textbf{0.02$\pm$0.00}    & 0.65$\pm$0.28    & \underline{0.09$\pm$0.07} & \textbf{0.00$\pm$0.00}    & \underline{0.03$\pm$0.00} & 0.18$\pm$0.00             & 0.13$\pm$0.10             \\
      Sarcos                    & 0.95$\pm$0.00                        & \textbf{-0.33$\pm$0.01}               & 1.41$\pm$0.41                        & \underline{-0.15$\pm$0.07} & 0.26$\pm$0.00             & \textbf{0.11$\pm$0.00}    & 0.45$\pm$0.21    & \underline{0.12$\pm$0.01} & 0.17$\pm$0.01             & \textbf{0.16$\pm$0.00}    & 0.23$\pm$0.01             & \underline{0.17$\pm$0.02} \\
      California housing        & 1.04$\pm$0.00                        & \textbf{0.56$\pm$0.03}                & 2.50$\pm$0.59                        & \underline{0.75$\pm$0.05}  & 0.32$\pm$0.00             & \textbf{0.25$\pm$0.01}    & 1.75$\pm$1.19    & \underline{0.26$\pm$0.02} & \textbf{0.35$\pm$0.00}    & \underline{0.35$\pm$0.01} & 0.35$\pm$0.00             & 0.36$\pm$0.03             \\
      Cpu activity              & 0.92$\pm$0.00                        & \textbf{-0.69$\pm$0.03}               & 1.75$\pm$0.65                        & \underline{-0.60$\pm$0.06} & 0.24$\pm$0.00             & \underline{0.07$\pm$0.00} & 0.70$\pm$0.40    & \textbf{0.07$\pm$0.00}    & 0.10$\pm$0.00             & 0.10$\pm$0.00             & \underline{0.10$\pm$0.00} & \textbf{0.10$\pm$0.00}    \\
      Diamonds                  & 0.95$\pm$0.00                        & \textbf{-1.11$\pm$0.03}               & 0.58$\pm$0.17                        & \underline{-0.85$\pm$0.24} & 0.26$\pm$0.00             & \textbf{0.11$\pm$0.00}    & 0.20$\pm$0.03    & \underline{0.11$\pm$0.01} & \underline{0.16$\pm$0.00} & \textbf{0.16$\pm$0.00}    & 0.18$\pm$0.00             & 0.16$\pm$0.01             \\
      Kin8nm                    & 1.20$\pm$0.00                        & \textbf{1.08$\pm$0.00}                & 2.61$\pm$2.59                        & \underline{1.15$\pm$0.10}  & \underline{0.43$\pm$0.00} & \textbf{0.42$\pm$0.00}    & 0.52$\pm$0.04    & 0.45$\pm$0.04             & 0.60$\pm$0.00             & \underline{0.60$\pm$0.00} & \textbf{0.60$\pm$0.00}    & 0.64$\pm$0.06             \\
      Pumadyn32nh               & 1.36$\pm$0.01                        & \underline{1.27$\pm$0.02}             & 1.99$\pm$0.26                        & \textbf{1.26$\pm$0.08}     & 0.53$\pm$0.00             & \textbf{0.51$\pm$0.01}    & 0.82$\pm$0.18    & \underline{0.52$\pm$0.03} & 0.74$\pm$0.01             & \underline{0.74$\pm$0.01} & \textbf{0.73$\pm$0.00}    & 0.76$\pm$0.03             \\
      Miami housing             & 0.99$\pm$0.00                        & \textbf{-0.46$\pm$0.01}               & 1.56$\pm$1.38                        & \underline{-0.29$\pm$0.08} & 0.28$\pm$0.00             & \textbf{0.17$\pm$0.00}    & 0.32$\pm$0.05    & \underline{0.18$\pm$0.01} & \textbf{0.24$\pm$0.01}    & \underline{0.25$\pm$0.00} & 0.28$\pm$0.01             & 0.26$\pm$0.01             \\
      Cps88wages                & \underline{1.22$\pm$0.00}            & \textbf{1.13$\pm$0.05}                & 5.84$\pm$10.01                       & 1.41$\pm$0.07              & 0.41$\pm$0.00             & \underline{0.39$\pm$0.01} & 0.48$\pm$0.02    & \textbf{0.37$\pm$0.01}    & 0.51$\pm$0.01             & 0.51$\pm$0.01             & \textbf{0.51$\pm$0.00}    & \underline{0.51$\pm$0.02} \\
      Kings county              & 1.01$\pm$0.00                        & \underline{0.13$\pm$0.01}             & 1.93$\pm$0.34                        & \textbf{0.04$\pm$0.07}     & 0.29$\pm$0.00             & \underline{0.19$\pm$0.00} & 0.87$\pm$0.35    & \textbf{0.17$\pm$0.01}    & 0.25$\pm$0.01             & 0.26$\pm$0.00             & \underline{0.24$\pm$0.00} & \textbf{0.24$\pm$0.01}    \\
      Brazilian houses          & \textbf{1.14$\pm$0.00}               & 33.28$\pm$8.68                        & \underline{11.15$\pm$16.45}          & 59.34$\pm$14.56            & 0.27$\pm$0.00             & \underline{0.09$\pm$0.00} & 0.20$\pm$0.03    & \textbf{0.08$\pm$0.00}    & \underline{0.11$\pm$0.00} & 0.12$\pm$0.00             & 0.12$\pm$0.00             & \textbf{0.10$\pm$0.00}    \\
      Health insurance          & \underline{1.24$\pm$0.00}            & \textbf{1.12$\pm$0.00}                & 1.99$\pm$0.57                        & 6.56$\pm$2.68              & 0.46$\pm$0.00             & \textbf{0.44$\pm$0.00}    & 0.58$\pm$0.07    & \underline{0.45$\pm$0.01} & \textbf{0.65$\pm$0.00}    & 0.65$\pm$0.00             & \underline{0.65$\pm$0.00} & 0.65$\pm$0.01             \\
      Fifa                      & \underline{1.07$\pm$0.03}            & \textbf{-0.33$\pm$0.12}               & 1.26$\pm$0.28                        & 6.32$\pm$10.44             & 0.31$\pm$0.01             & \textbf{0.16$\pm$0.00}    & 0.48$\pm$0.12    & \underline{0.19$\pm$0.03} & 0.25$\pm$0.02             & \textbf{0.22$\pm$0.00}    & \underline{0.23$\pm$0.01} & 0.25$\pm$0.03             \\
      Space ga                  & 1.05$\pm$0.01                        & \textbf{0.59$\pm$0.01}                & 4.45$\pm$0.76                        & \underline{0.73$\pm$0.15}  & 0.32$\pm$0.00             & \textbf{0.25$\pm$0.00}    & 14.92$\pm$6.59   & \underline{0.28$\pm$0.05} & 0.37$\pm$0.01             & \textbf{0.35$\pm$0.00}    & \underline{0.36$\pm$0.00} & 0.40$\pm$0.07             \\
      \midrule
      Average rank $\downarrow$ & 2.71                                 & \textbf{1.17}                         & 3.75                                 & \underline{2.38}           & 2.88                      & \textbf{1.29}             & 3.92             & \underline{1.92}          & 2.54                      & \underline{2.33}          & \textbf{2.29}             & 2.83                      \\
      \bottomrule
    \end{tabular}}
  \label{tab:predictive_performance_regression}
\end{table*}

We use a similar setting for classification, with 10 real-world datasets from the OpenML suite 99~\parencite{bischl.2021}. We selected \ac{NAMLSS} -- \ac{NAM} and \ac{NAMLSS} become the same when applied to classification -- and \emph{EnsNAM} as baselines. Similar to the related work, we report:
\begin{enumerate*}[label=(\arabic*)]
  \item Accuracy,
  \item Brier score,
  \item AUROC, and
  \item Calibration.
\end{enumerate*}
The results are shown in Table~\ref{tab:predictive_performance_classification}. On average, {\evinam} trails the baselines but remains competitive, including best and second-best scores on kr-vs-kp and Churn. Unlike \ac{NAMLSS} and EnsNAM, {\evinam} provides both aleatoric and epistemic uncertainty in a single pass, preserves feature-additive class probabilities (in contrast to \ac{NAMLSS} and \ac{NAM}), and avoids the computational overhead of EnsNAM.

\begin{table*}[t]
  \centering
  \caption{Classification results on test data via mean$\pm$standard deviation; average rank in last row.}
  \resizebox{\textwidth}{!}{
    \begin{tabular}{@{}lccc|ccc|ccc|ccc@{}}
      \toprule
                                & \multicolumn{3}{c}{Accuracy $\uparrow$} & \multicolumn{3}{c}{Brier score $\downarrow$} & \multicolumn{3}{c}{AUROC $\uparrow$} & \multicolumn{3}{c}{Calibration $\downarrow$}                                                                                                                                                                                                                              \\
      Dataset                   & NAMLSS                                  & EnsNAM                                       & \textbf{{\evinam}}                   & NAMLSS                                       & EnsNAM                    & \textbf{{\evinam}}     & NAMLSS                    & EnsNAM                    & \textbf{{\evinam}}        & NAMLSS                    & EnsNAM                    & \textbf{{\evinam}}        \\ \midrule
      Kr-vs-kp                  & 0.96$\pm$0.01                           & \textbf{0.97$\pm$0.00}                       & \underline{0.97$\pm$0.00}            & 0.06$\pm$0.01                                & \underline{0.05$\pm$0.00} & \textbf{0.05$\pm$0.00} & 1.00$\pm$0.00             & \underline{1.00$\pm$0.00} & \textbf{1.00$\pm$0.00}    & \textbf{0.02$\pm$0.00}    & \underline{0.02$\pm$0.00} & 0.03$\pm$0.00             \\
      Letter                    & \underline{0.85$\pm$0.01}               & \textbf{0.88$\pm$0.00}                       & 0.48$\pm$0.00                        & \underline{0.21$\pm$0.01}                    & \textbf{0.18$\pm$0.00}    & 0.68$\pm$0.01          & \underline{0.99$\pm$0.00} & \textbf{1.00$\pm$0.00}    & 0.93$\pm$0.00             & \textbf{0.02$\pm$0.00}    & \underline{0.02$\pm$0.00} & 0.13$\pm$0.00             \\
      Mfeat-fourier             & \underline{0.83$\pm$0.02}               & \textbf{0.85$\pm$0.01}                       & 0.61$\pm$0.01                        & \underline{0.27$\pm$0.03}                    & \textbf{0.22$\pm$0.02}    & 0.50$\pm$0.02          & \underline{0.98$\pm$0.00} & \textbf{0.99$\pm$0.00}    & 0.93$\pm$0.01             & 0.11$\pm$0.02             & \textbf{0.07$\pm$0.02}    & \underline{0.08$\pm$0.02} \\
      Mfeat-karhunen            & \textbf{0.94$\pm$0.01}                  & \underline{0.93$\pm$0.01}                    & 0.63$\pm$0.01                        & \textbf{0.10$\pm$0.01}                       & \underline{0.11$\pm$0.01} & 0.52$\pm$0.01          & \textbf{1.00$\pm$0.00}    & \underline{1.00$\pm$0.00} & 0.93$\pm$0.01             & \underline{0.04$\pm$0.01} & \textbf{0.04$\pm$0.02}    & 0.12$\pm$0.01             \\
      Mfeat-morphological       & \underline{0.71$\pm$0.02}               & \textbf{0.73$\pm$0.01}                       & 0.68$\pm$0.01                        & \textbf{0.37$\pm$0.00}                       & \underline{0.37$\pm$0.01} & 0.44$\pm$0.01          & \underline{0.96$\pm$0.00} & \textbf{0.96$\pm$0.00}    & 0.93$\pm$0.00             & \underline{0.08$\pm$0.01} & \textbf{0.06$\pm$0.01}    & 0.09$\pm$0.01             \\
      Mfeat-zernike             & \underline{0.78$\pm$0.02}               & \textbf{0.80$\pm$0.01}                       & 0.62$\pm$0.01                        & \underline{0.28$\pm$0.02}                    & \textbf{0.26$\pm$0.01}    & 0.51$\pm$0.01          & \underline{0.97$\pm$0.00} & \textbf{0.97$\pm$0.00}    & 0.93$\pm$0.00             & \underline{0.09$\pm$0.01} & \textbf{0.07$\pm$0.01}    & 0.10$\pm$0.02             \\
      Cmc                       & \underline{0.60$\pm$0.01}               & \textbf{0.61$\pm$0.02}                       & 0.59$\pm$0.01                        & \textbf{0.51$\pm$0.01}                       & \underline{0.52$\pm$0.01} & 0.52$\pm$0.00          & \textbf{0.77$\pm$0.01}    & 0.77$\pm$0.01             & \underline{0.77$\pm$0.00} & \textbf{0.07$\pm$0.02}    & 0.09$\pm$0.01             & \underline{0.07$\pm$0.01} \\
      Optdigits                 & \textbf{0.97$\pm$0.00}                  & \underline{0.97$\pm$0.00}                    & 0.68$\pm$0.01                        & \textbf{0.05$\pm$0.00}                       & \underline{0.06$\pm$0.00} & 0.48$\pm$0.00          & \underline{1.00$\pm$0.00} & \textbf{1.00$\pm$0.00}    & 0.95$\pm$0.00             & \textbf{0.02$\pm$0.00}    & \underline{0.03$\pm$0.00} & 0.10$\pm$0.01             \\
      Pendigits                 & \underline{0.98$\pm$0.00}               & \textbf{0.99$\pm$0.00}                       & 0.66$\pm$0.02                        & \underline{0.04$\pm$0.00}                    & \textbf{0.02$\pm$0.00}    & 0.46$\pm$0.02          & \underline{1.00$\pm$0.00} & \textbf{1.00$\pm$0.00}    & 0.95$\pm$0.00             & \underline{0.01$\pm$0.00} & \textbf{0.01$\pm$0.00}    & 0.06$\pm$0.02             \\
      Churn                     & \textbf{0.91$\pm$0.00}                  & 0.89$\pm$0.00                                & \underline{0.91$\pm$0.00}            & \underline{0.15$\pm$0.00}                    & 0.17$\pm$0.00             & \textbf{0.14$\pm$0.00} & \textbf{0.86$\pm$0.01}    & 0.83$\pm$0.02             & \underline{0.84$\pm$0.01} & 0.04$\pm$0.01             & \underline{0.04$\pm$0.01} & \textbf{0.03$\pm$0.01}    \\
      \midrule
      Average rank $\downarrow$ & \underline{1.80}                        & \textbf{1.40}                                & 2.80                                 & \textbf{1.70}                                & \textbf{1.70}             & 2.60                   & \underline{1.80}          & \textbf{1.60}             & 2.60                      & \underline{1.80}          & \textbf{1.60}             & 2.60                      \\
      \bottomrule
    \end{tabular}}
  \label{tab:predictive_performance_classification}
\end{table*}

We do not include $\text{DER}_{\text{MLP}}$ in the tables because $\text{DER}_{\text{MLP}}$ depends on feature interactions, making it outside the \ac{NAM}-style hypothesis class targeted by {\evinam}. As expected, on average over the first five datasets for regression, $\text{DER}_{\text{MLP}}$ outperformed {\evinam}, e.g., with an \ac{MAE} of 0.23 compared to 0.49. Similarly, for classification, $\text{DER}_{\text{MLP}}$ achieved an accuracy of 0.90 compared to 0.71.

\paragraph{Computational Complexity} Averaged over 10 models and the first 10 tasks of OpenML suites 353 (regression) and 99 (classification) per 100 epochs, the training times for regression were 82s for \ac{NAM}, 106s for \ac{NAMLSS}, 120s for {\evinam}, and 342s for EnsNAM. For classification, the corresponding values were 46s for \ac{NAMLSS}, 48s for {\evinam}, and 155s for EnsNAM. As expected, {\evinam} is slightly slower than \ac{NAM} and \ac{NAMLSS} in regression -- it learns a richer set of distributional parameters -- but faster than EnsNAM. For classification the training time of {\evinam} is comparable to that of \ac{NAMLSS}. Compared to EnsNAM, {\evinam} uses less memory and compute for both training and inference. {\evinam} computes feature contributions and both uncertainties more efficiently than \acp{BNN} or BNAM, as it does not rely on sampling.

\begin{figure}
  \centering
  \begin{minipage}[t]{0.48\linewidth}
    \centering
    \includegraphics[width=\linewidth]{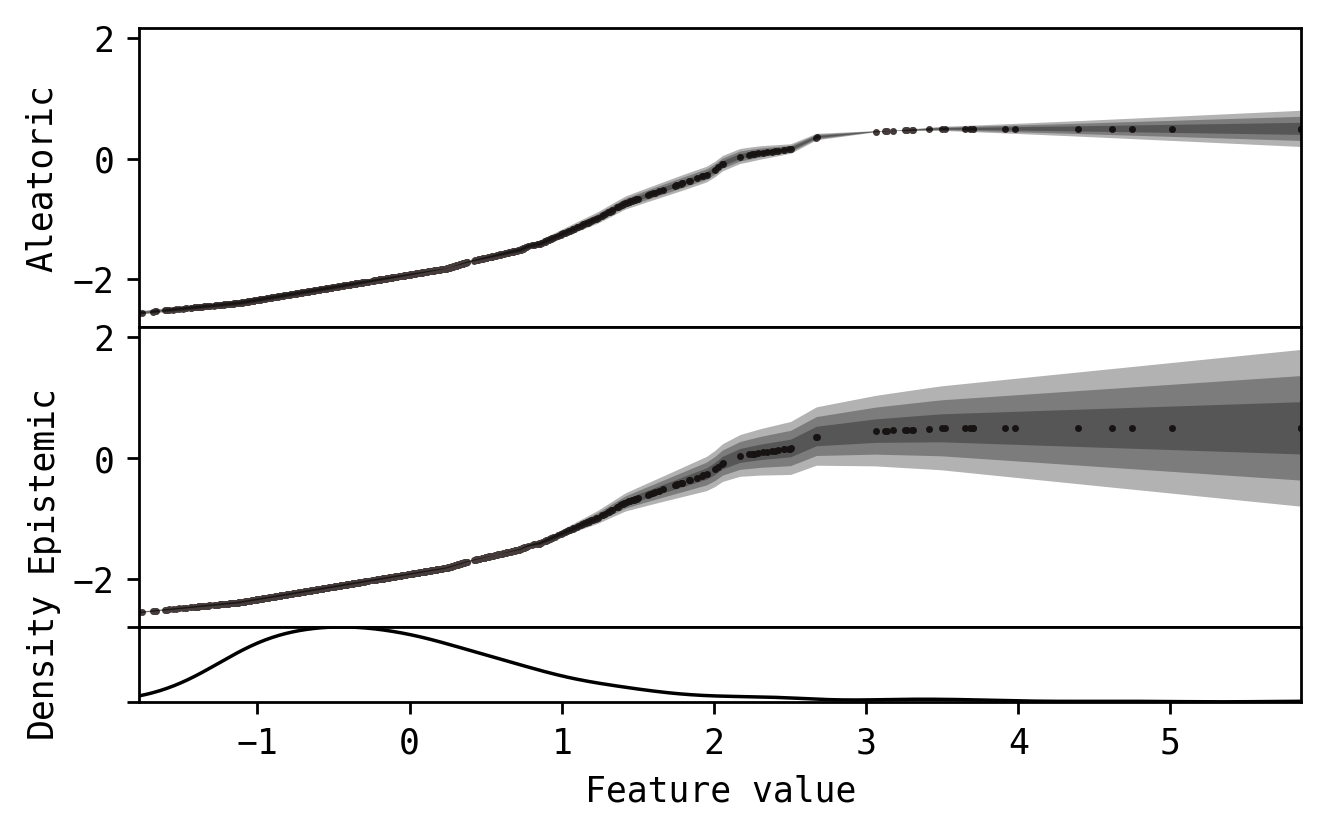}
    \caption*{(a) Feature \emph{median income}}
  \end{minipage}
  \begin{minipage}[t]{0.48\linewidth}
    \centering
    \includegraphics[width=\linewidth]{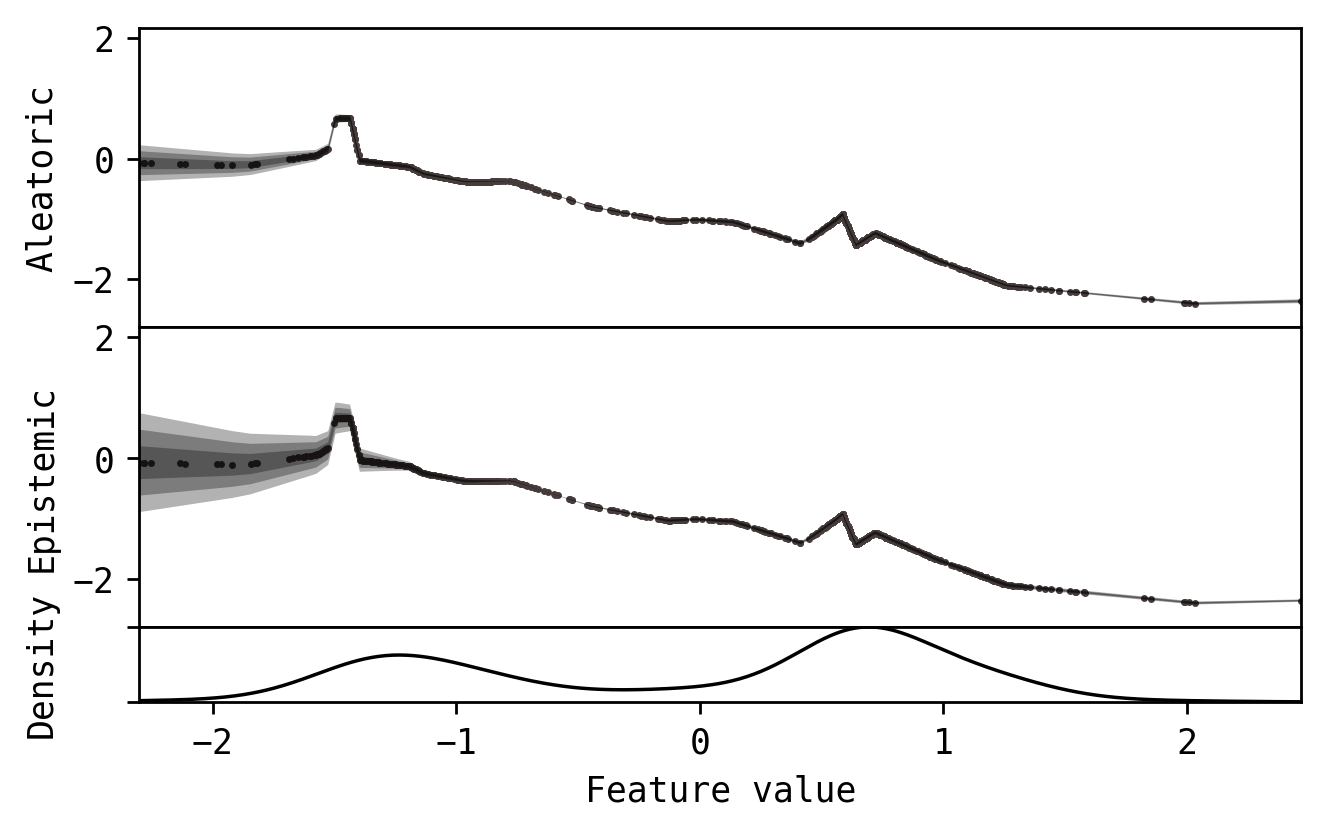}
    \caption*{(b) Feature \emph{longitude}}
  \end{minipage}
  \caption{Contribution of two features to the prediction (black dots) along with the contribution of each feature to the uncertainties (gray shadows) as well as the density distribution of the feature values; uncertainty values are smoothed via LOWESS -- see~\cite{cleveland.1979}.}
  \label{fig:FeatureContributionAndUncertainties}
\end{figure}

\paragraph{Intelligibility} We selected the California housing dataset and one of the models previously trained. In Figure~\ref{fig:FeatureContributionAndUncertainties} shows the contribution of two features (\emph{median income} and \emph{longitude}) to the predicted house price along with the both uncertainties. As shown, {\evinam} learned that an increase in income leads to a higher house price. Further, the aleatoric uncertainty was higher in areas with fewer observations. The model revealed that the epistemic uncertainty was higher for higher median income values. For the longitude feature, both uncertainties correlated, indicating that uncertainty was prevalent in the data and the model's prediction.

\section{Conclusion}
In this work, we presented {\evinam}, an extension of \ac{DER} that unites the intelligibility of \acp{NAM} -- enabled by its additive structure -- with the principled uncertainty estimation via evidential learning. Unlike existing approaches such as \ac{BNN}s, \ac{NAMLSS}, or vanilla \ac{DER}, {\evinam} offers transparent, single-pass estimation (at inference time) of aleatoric and epistemic uncertainty, as well as explicit feature contributions to the prediction. In addition to the integration of \ac{DER} and \ac{NAM}, this is made possible by forwarding the nonlinearity needed to satisfy distributional requirements.

Our experimental results on synthetic and real-world datasets reveal that {\evinam} performed on par with \ac{NAM}, EnsNAM, and \ac{NAMLSS} in terms of predictive performance for regression, while uniquely providing both aleatoric and epistemic uncertainties and per-feature contributions. Unlike \ac{NAM} or \ac{NAMLSS}, for classification, {\evinam} provides feature-additive class probabilities. Despite critiques regarding the estimation of epistemic uncertainty, evidential learning is useful in settings that can defer to humans as it rises.

Our approach can be extended to \acp{GAM} and to those additive models with pairwise interactions. Also, forwarding the nonlinearity is independent of the activation function and applies to \ac{NAM} and \ac{NAMLSS}, offering the opportunity of additive class probabilities.

Making the uncertainties explicit allows human decision-makers to identify out-of-domain samples and assess a model's confidence in predictions. These capabilities mark an important step toward more interpretable and trustworthy models.

\printbibliography[heading=bibintoc]

\end{document}